\documentclass[runningheads]{llncs}
\usepackage[T1]{fontenc}
\usepackage{graphicx}
\usepackage{hyperref}
\usepackage{fontawesome5}
\makeatletter
\newcommand{\github}[1]{%
   \href{#1}{\faGithubSquare}%
}
\makeatother

\begin{document}
\title{Extending Token Computation for LLM Reasoning}
%
%\titlerunning{Abbreviated paper title}
% If the paper title is too long for the running head, you can set
% an abbreviated paper title here
%
\author{Liao Bingli \and
Danilo Vasconcellos Vargas}
\authorrunning{B. Liao et al.}
% First names are abbreviated in the running head.
% If there are more than two authors, 'et al.' is used.
%
\institute{Kyushu University, Fukuoka, Japan
\email{\{liao.bingli.734@s, vargas@inf\}.kyushu-u.ac.jp}}
%

% \author{anonymous}

\maketitle              % typeset the header of the contribution
\begin{abstract}
Large Language Models (LLMs) are pivotal in advancing natural language processing but often struggle with complex reasoning tasks due to inefficient attention distributions. In this paper, we explore the effect of increased computed tokens on LLM performance and introduce a novel method for extending computed tokens in the Chain-of-Thought (CoT) process, utilizing attention mechanism optimization. By fine-tuning an LLM on a domain-specific, highly structured dataset, we analyze attention patterns across layers, identifying inefficiencies caused by non-semantic tokens with outlier high attention scores. To address this, we propose an algorithm that emulates early layer attention patterns across downstream layers to re-balance skewed attention distributions and enhance knowledge abstraction. Our findings demonstrate that our approach not only facilitates a deeper understanding of the internal dynamics of LLMs but also significantly improves their reasoning capabilities, particularly in non-STEM domains. Our study lays the groundwork for further innovations in LLM design, aiming to create more powerful, versatile, and responsible models capable of tackling a broad range of real-world applications.
\href{https://github.com/metacarbon/attentionReasoning-llm}{\faGithubSquare}   
\end{abstract}
\section{Introduction}
Large Language Models (LLMs) have demonstrated exceptional capabilities in natural language understanding, generation, and reasoning, thanks to advancements by \cite{radford2018improving,openai2023gpt,roziere2023code,katz2024gpt,li2024first}. Recent innovations have focused on enhancing the performance of these models through various techniques including Chain-of-Thought (CoT) methods \cite{wei2022chain}, fine-tuning approaches \cite{hu2021lora}, and Retrieval-Augmented Generation (RAG) \cite{lewis2020retrieval}. The CoT, in particular, suggests that extending the computation of more tokens before generating the final answer could significantly improve reasoning abilities. Our study explores whether an extended CoT chain, within the constraints of the same prompt, can achieve superior reasoning outcomes.

Attention mechanisms in LLMs play a pivotal role, especially in processing long sequences \cite{xiao2023efficient} and in scenarios involving massive activation \cite{sun2024massive}. Despite these insights, there is scant research on harnessing these mechanisms to fundamentally boost the reasoning capabilities of LLMs. Our research aims to bridge this gap by investigating the inner workings of attention mechanisms and proposing a novel algorithm that optimizes attention patterns. This algorithm allows for the extension of computed tokens without altering the prompt or requiring additional training, inspired by the principles of in-context learning \cite{brown2020language} and CoT.

To facilitate a focused analysis of LLMs' internal mechanisms, we fine-tune a model using domain-specific, highly structured multi-conversation data. Our observations reveal massive activation across model layers, corroborating findings by \cite{sun2024massive}. However, our further exploration indicates that this high activation skews attention weight distributions, potentially obstructing the full realization of the model's capabilities.

Building on these insights, we have developed and implemented a novel algorithm within the LLaMA framework \cite{touvron2023LLaMA}, which compensates for the observed loss of attention across layers. We demonstrate that our approach not only increases the number of tokens computed under the same prompt constraints but also enhances reasoning abilities, particularly in non-STEM domains, and improves instruction-following capabilities. Our findings highlight the critical impact of computed token numbers on the reasoning abilities of LLMs.

\begin{figure}[]
  \centering
  \includegraphics[width=0.85\textwidth]{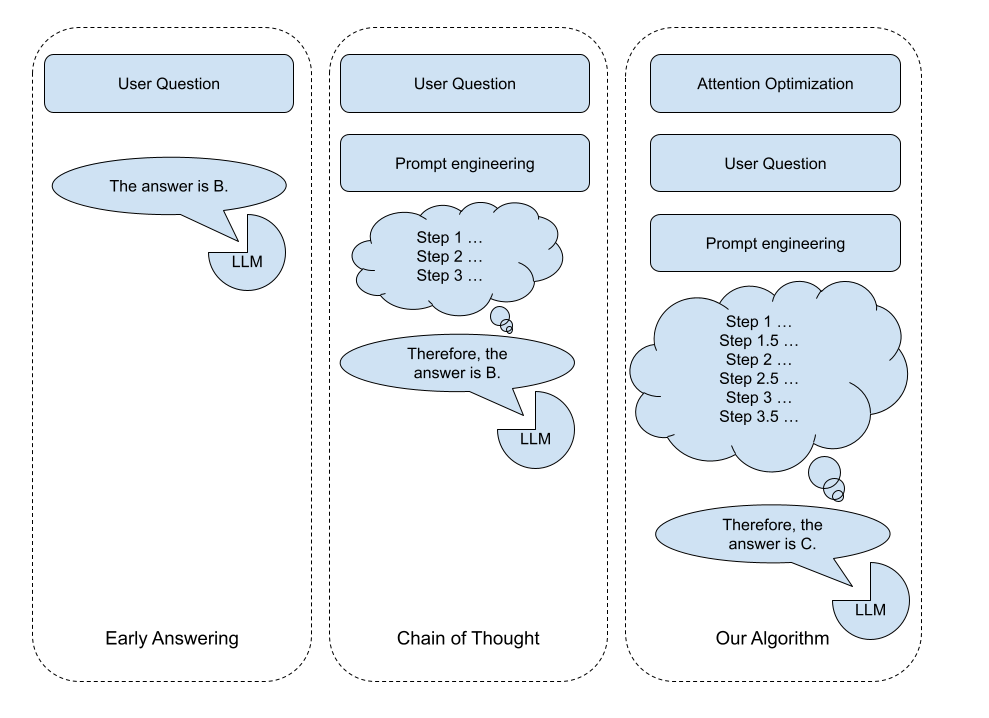}
  \caption{Illustration of the big picture of our method.}
  \label{fig:bigpic}
\end{figure}

\section{Related Work}
Attention mechanisms are a foundational component of LLMs and have been extensively studied. Early works like those by \cite{adi2016fine}, who proposed a probing method to elucidate vector representations, and \cite{yeh2023attentionviz}, who introduced joint embeddings to enhance the interpretability of attention visualizations, provided significant insights for small neural networks and simple input sentences. However, these studies fall short of offering a comprehensive understanding of attention dynamics across different layers in LLMs.

\cite{clark2019does} demonstrated that BERT models often capture syntactic relationships by focusing on delimiter tokens, and \cite{xiao2023efficient} highlighted how beginning tokens in long sequences can compress positional information using window attention. These studies, however, primarily analyze outlier high activation patterns and overlook the broader spectrum of attention patterns that, despite their less extreme activation, are crucial in the generation process.

\cite{sun2024massive} observed massive activations in LLM hidden states leading to concentrated attention. They attempted to mitigate these effects by introducing additional learnable weights to balance attention. Nevertheless, their approach focuses excessively on these high-activation outliers, neglecting the potential impact on other tokens, which may degrade the overall reasoning capability of self-attention-based models.

Building on these foundations, our work advances the interpretability and implementation of attention patterns within LLMs. By examining attention mechanisms across layers, validated by standard benchmarks, we aim to deliver a more nuanced understanding of the internal mechanics of LLMs. Our algorithm, which extends the number of computed tokens without modifying the initial prompt, demonstrates enhanced reasoning capabilities through an extended CoT.

In the context of training LMs to reason, our work differs from approaches that rely on mined reasoning traces or reasoning-like data \cite{chung2022scaling,lewkowycz2022solving}, which can be sensitive to annotator capability, expensive, and difficult to scale. Instead, we leverage the LM's own generated reasoning, building on the literature on self-play \cite{polu2020generative} and extending the ideas of the Self-Taught Reasoner \cite{zelikman2022star}.

\begin{figure}[]
  \centering
  \includegraphics[width=1.0\textwidth]{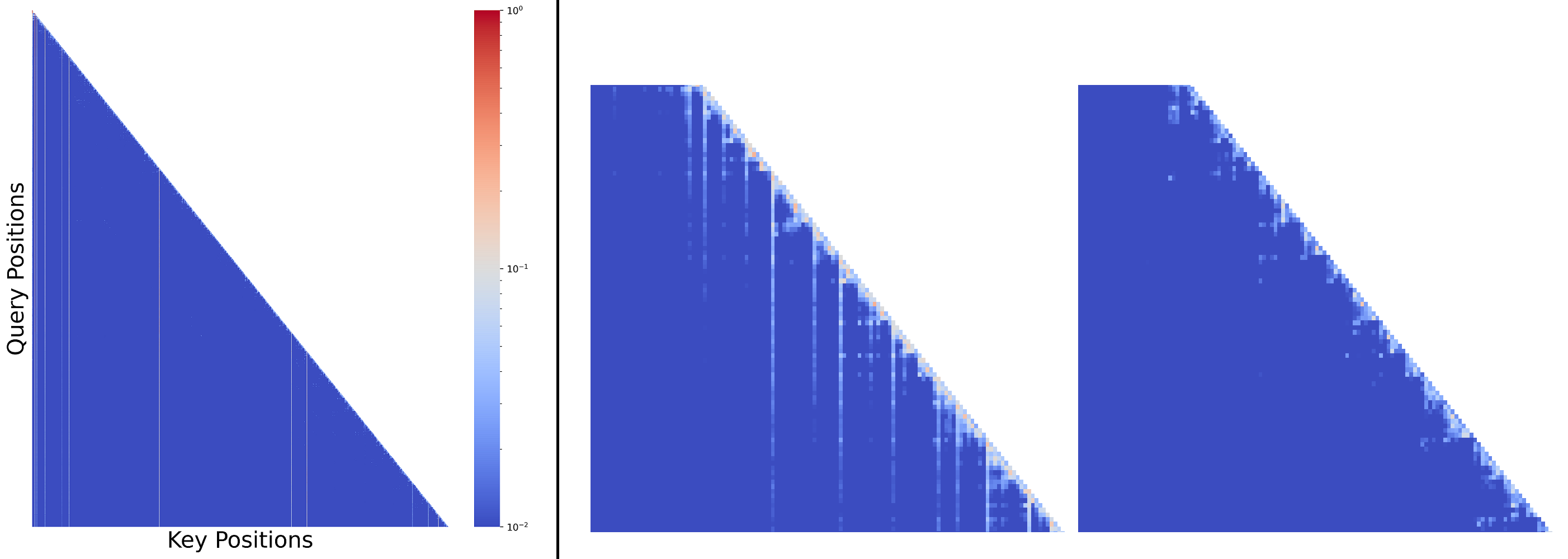}
  \caption{Visualization of attention score matrices from the fine-tuned language model. (Left) Attention score matrix from layer 6 of the fine-tuned language model, providing an overview of the attention patterns learned at this depth. (Right) Zoomed-in view of attention scores from the first and middle layers of the model, offering a more detailed view of the attention dynamics within these layers.}
  \label{fig:att_layer6}
\end{figure}

\section{Motivations}
The complexity of publicly available LLMs poses significant challenges for understanding their internal mechanisms. These models, typically trained on vast corpora, result in intricate inner weights that obscure their operational principles. Moreover, re-training LLMs with the original datasets after algorithmic modifications is computationally costly and time-consuming. To address these challenges, we opted to fine-tune an LLM using a domain-specific, prompt-based, multi-turn conversation dataset \cite{liao2023towards}. This approach not only reduces the model's complexity but also facilitates a more focused investigation of its functionalities.

Our primary focus was on exploring the behavior of attention patterns within a fine-tuned 7B LLM. We analyzed the average multi-head attention scores across each layer, as illustrated in Figure \ref{fig:att_layer6}. The observed patterns, characterized by vertical lines in the attention maps, suggest focused attention on specific tokens—referred to as "anchor tokens". This observation is partially consistent with previous studies that emphasize the significance of initial tokens in attention mechanisms \cite{xiao2023efficient}.

\begin{figure}[ht]
  \centering
  \includegraphics[width=1.0\textwidth]{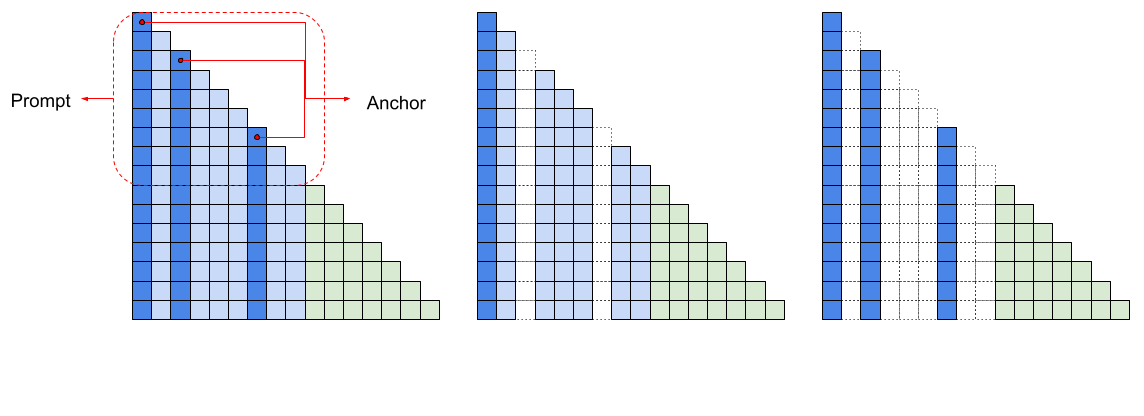}
  \caption{Illustration of attention score matrix modifications in preliminary tests on the fine-tuned language model. The attention score matrix is segmented into the prompt and dialogue parts. (Left) The original attention score matrix, with dark blue cells representing anchor tokens. (Middle) Modified attention score matrix with anchor tokens removed from the prompt-related token range. (Right) Counter-experiment where only anchor tokens are retained in the prompt-related token range.}
  \label{fig:att_adjust}
\end{figure}

To investigate the role of these anchor tokens, as shown in Figure \ref{fig:att_adjust}, we conducted a series of controlled experiments:
\begin{enumerate}
    \item \textbf{Isolating Anchor Tokens:} We manually adjusted the attention scores by setting non-anchor tokens in the prompt part of middle layers (4 to 8) to zero, while retaining those of the anchor tokens. This experiment demonstrated that the LLM could still generate high-fidelity, multi-turn conversations under these modified conditions.
    \item \textbf{Removing Anchor Tokens:} Conversely, we experimented with removing the attention scores attributed to the anchor tokens, which resulted in a performance similar to the first experiment. This suggests that while anchor tokens are influential, they are not solely responsible for the generation capabilities.
    \item \textbf{Testing Recent Time-Step Tokens:} We reset the attention scores of the most recent time-step tokens, resulting in a perplexity explosion, highlighting their critical role in knowledge abstraction.
\end{enumerate}

In further tests, we eliminated attention scores entirely in every alternate layer (layers 4 to 8), allowing the Feed-Forward Network (FFN) to process only newly generated tokens, with information from previous tokens carried over through residual connections. The LLM maintained its performance, suggesting that middle layers can operate effectively even with reduced direct attention cues. However, similar modifications in the initial layers (1 and 2) or completely bypassing any Attention-FFN layers resulted in significant performance degradation, underscoring the importance of early layers in initial information processing and integration.

\begin{figure}[ht]
  \centering
  \includegraphics[width=1.0\textwidth]{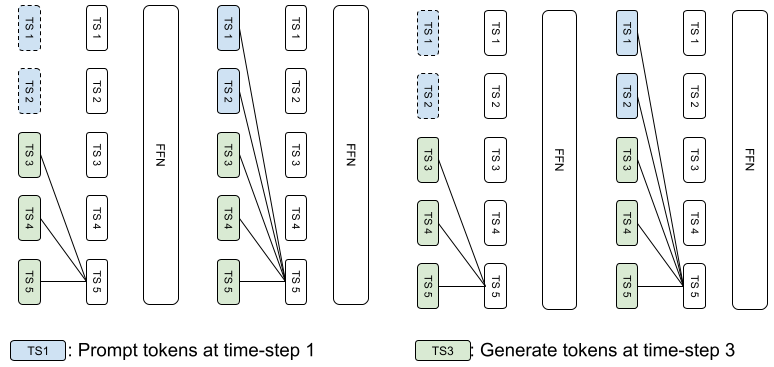}
  \caption{The figure illustrates the experiment where the prompt range attention scores were removed from alternating middle layers (4 to 8).}
  \label{fig:alterSeenRange}
\end{figure}

\section{Attention Mechanism Modification}
\subsection{Insights into Attention Patterns}
Our analysis of attention patterns within the fine-tuned LLM uncovered that highly activated attention in the middle layers disproportionately focuses on non-semantic tokens such as "is", "the", and newline characters. This observation echoes findings by \cite{sun2024massive}, where such tokens, commonly frequent in English texts, draw significant attention. According to the Corpus of Contemporary American English (COCA) \cite{davies2010corpus}, tokens like "the" and "is" are among the most common in English, ranking first and second respectively. Although COCA does not account for newline tokens, they similarly exhibit high frequency in our fine-tuning dataset. These tokens absorb a large proportion of attention scores, leading to a skewed distribution within the self-attention mechanism.

\begin{figure}[ht]
  \centering
  \includegraphics[width=1.0\textwidth]{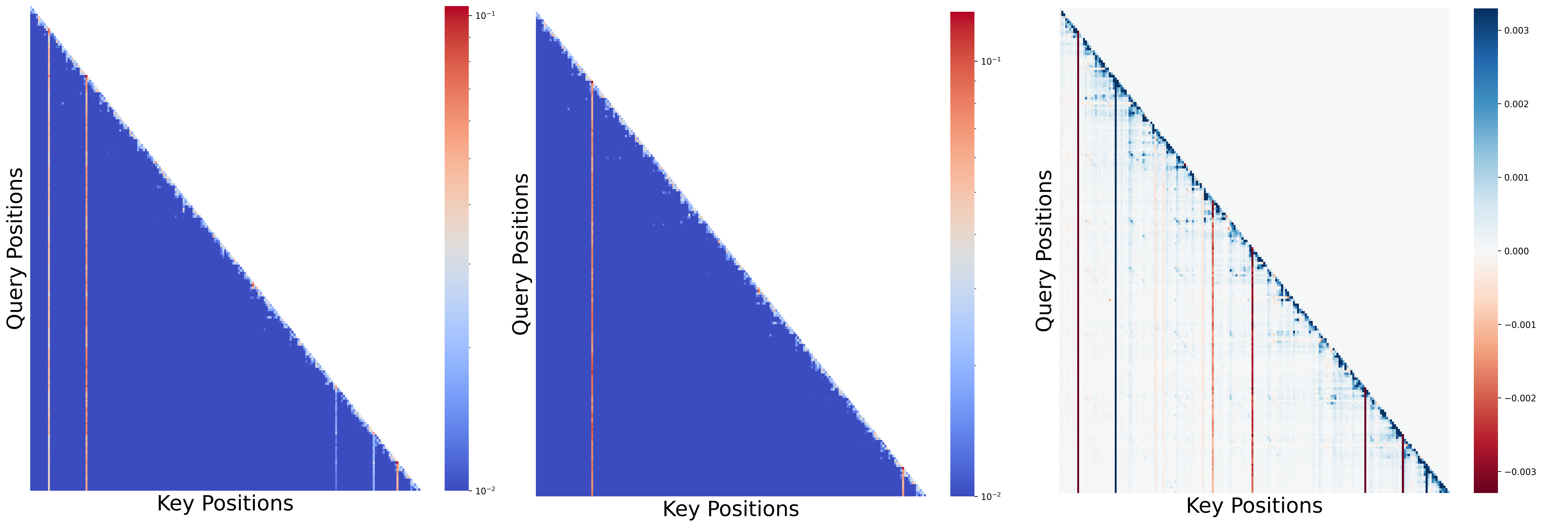}
  \caption{Comparative analysis of attention score matrices from the original fine-tuned LLM and the model fine-tuned with dropout. (Left) Attention score matrix from the original fine-tuned model. (Middle) Attention score matrix from the model fine-tuned with dropout. (Right) Matrix depicting the difference in attention scores between the original and dropout-regularized models, highlighting the impact of dropout on the learned attention patterns.}
  \label{fig:att_compare}
\end{figure}

To counteract this imbalance, we implemented a strategy of applying dropout to the multi-head attention output projection layer before it feeds into the FFN. This modification aims to recalibrate the attention weights, shifting focus towards semantically significant tokens that carry more meaningful information. Figure \ref{fig:att_compare} illustrates the effect of introducing a dropout rate of 0.1. This intervention significantly diminished the attention directed towards the previously dominant non-semantic tokens, as evidenced in the adjusted attention heatmaps.

The comparative analysis of attention heatmaps between an early layer and a middle layer, shown on the right side of Figure \ref{fig:att_layer6}, highlights a notable shift. In the early layers, semantic tokens effectively garner attention, directing information flow towards relevant latter time-steps. However, in middle layers, this focused attention is diluted by the high-score non-semantic tokens, which absorb attention away from semantically important previous time-step tokens.

\subsection{Proposed Algorithm for Attention Optimization}
Our analysis of attention patterns in LLMs has indicated a disparity between the attention distribution across layers: early layers maintain fine-grained attention on semantic tokens from previous time-steps, while attention in downstream layers becomes skewed, predominantly due to outlier tokens. This observation suggests that enhancing the retention of fine-grained attention in subsequent layers could potentially improve the model's performance. However, adjusting attention dynamically to emphasize important tokens from past time-steps remains challenging due to the complex interdependencies of attention patterns.

To address the skewed attention distribution in downstream layers, we developed an algorithm designed to emulate the attention pattern of the top layer across these layers, but with a moderating weight decay factor to prevent drastic disruptions to local attention dynamics. Rather than directly copying attention scores from the top layer, our algorithm amplifies the attention scores in downstream layers guided by the top layer's attention pattern. This approach avoids the complete replication of scores, which could overly homogenize attention across layers. Our proposed algorithm is formalized as follows:

Let \(A_l\) denote the attention matrix at layer \(l\), with \(A_l(i, j)\) representing the attention score from token \(i\) to token \(j\) at layer \(l\). Let \(M_t\) represent the top layer attention pattern mask, and \(h\) denote the maximum layer index (e.g., 31 in the case of LLaMA 7B). For each downstream layer \(l\), the attention matrix is updated according to the formula:

\[
A_l(i, j) = A_l(i, j) + A_l(i, j) \cdot \left(1 -  \frac{l}{h}\right) \cdot M_t(i, j), \quad \forall i, j \notin D
\]

where \(D\) represents the set of dialogical positions in the attention matrix, which are deliberately excluded from this adjustment to preserve the critical role of recent time-step tokens in knowledge abstraction.

This algorithm, applied without any additional training, aims to refocus the LLM's attention towards semantically important tokens by modifying its attention distribution, based on insights garnered from our prior empirical investigations.

\begin{figure}[]
  \centering
  \includegraphics[width=1.0\textwidth]{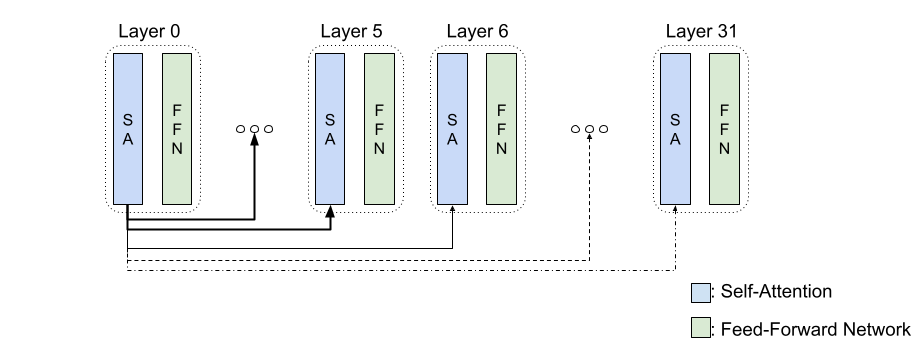}
  \caption{This figure illustrates the high-level architecture of our proposed algorithm. The key innovation lies in the hierarchical attention mechanism, where each downstream layer adaptively emphasizes its weight scores based on the attention patterns learned by the early layer. }
  \label{fig:algorithm}
\end{figure}

\section{Evaluation}
\subsection{Methodology}
To validate the effectiveness of our proposed algorithm, we utilized the LLaMA-2 model along with the MMLU, PIQA, and SIQA datasets as benchmarks \cite{hendrycks2020measuring,bisk2019piqa,sap2019socialiqa}. The LLaMA model serves as a robust baseline for evaluating the impact of our approach on enhancing reasoning capabilities. The MMLU dataset, covering a broad spectrum of academic subjects, is ideal for assessing the model's performance on complex and diverse reasoning tasks. The PIQA dataset specifically measures the model's reasoning in physical commonsense scenarios, while the SIQA dataset evaluates social commonsense intelligence.

We implemented our proposed attention pattern modification method on both the LLaMA 7B and 13B models. To assess the effectiveness of our algorithm, we conducted three distinct evaluations using a zero temperature setting: early answering, zero-shot CoT, and zero-shot CoT with our method applied. We hypothesize that the model's ability to accurately follow prompts with embedded reasoning steps is crucial for enhancing its overall reasoning performance. This is particularly pertinent in zero-shot CoT reasoning, which enables the model to address questions unresolvable through early answering by leveraging its reasoning capabilities.

Our experimental results indicate a decrease in accuracy for zero-shot CoT compared to early answering. This suggests that the LLM may recall question-answer pairs from its training and alignment phases, allowing it to respond directly from memory rather than engaging in multi-step reasoning. To ensure a more accurate evaluation, we isolated a subset containing about half of the dataset by filtering out questions that were solvable through early answering. This approach allowed us to focus on questions that remained unsolved through early answering but were potentially solvable using CoT reasoning. This methodology helps in distinguishing the true reasoning ability of the LLM from mere recall, providing a clearer assessment of our proposed algorithm.

\begin{figure}[]
  \centering
  \includegraphics[width=1.0\textwidth]{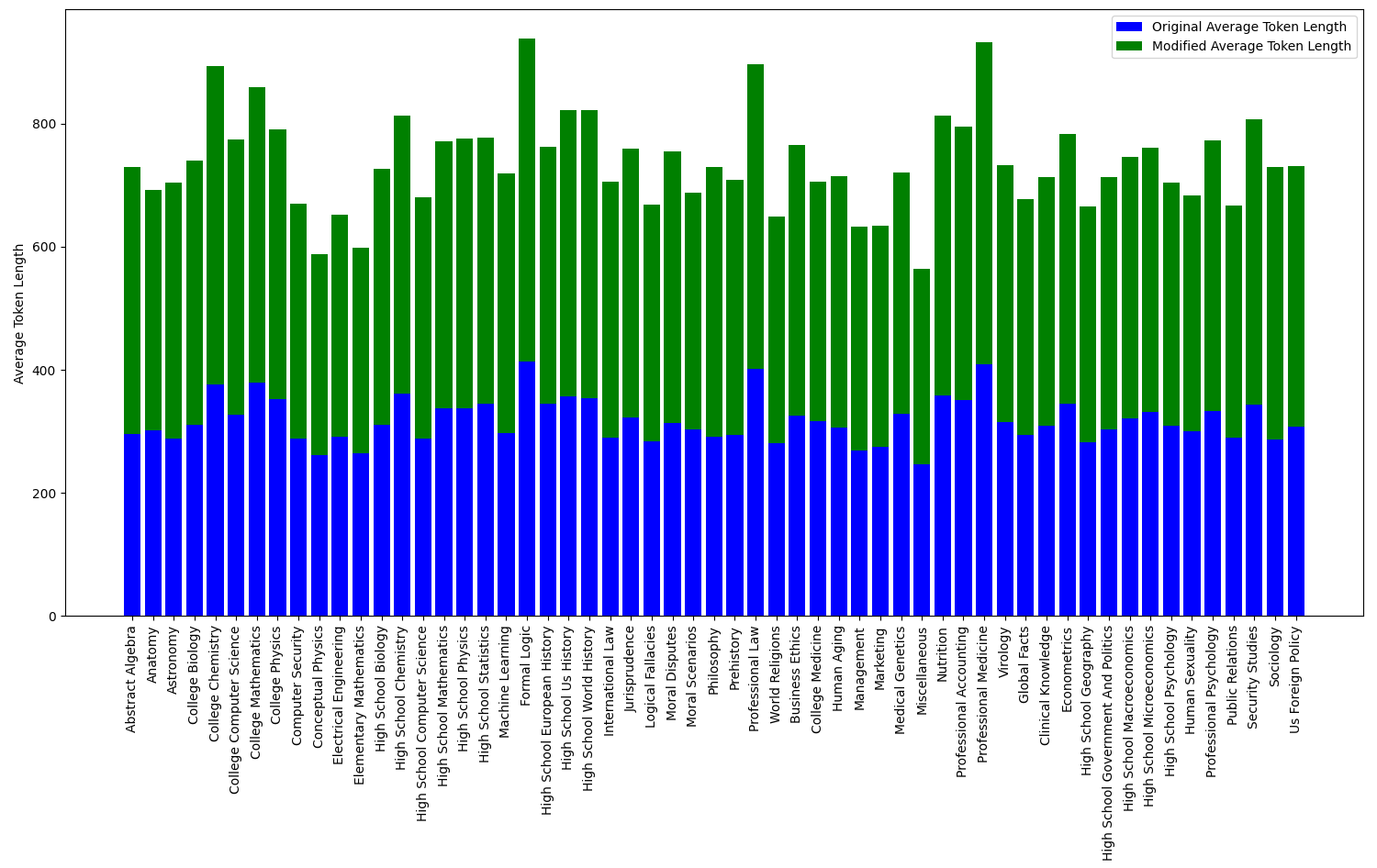}
  \caption{Comparative analysis of average token length required to solve questions using zero-shot CoT reasoning from the original LLaMA-2 model and LLaMA-2 enhanced with our proposed approach.}
  \label{fig:token_len}
\end{figure}

\subsection{Results and Discussion}
Table \ref{model-accuracy-table} provides a comparative analysis of the performance of our modified method against the baseline LLaMA model. For the 7B model, our method did not yield an improvement on the MMLU dataset. In contrast, the 13B model demonstrated significant enhancements in the Non-STEM categories of MMLU, as well as in the PIQA and SIQA datasets, suggesting that larger LLMs exhibit stronger CoT reasoning capabilities with our proposed algorithm.

Our hypothesis posited that the number of tokens calculated during reasoning impacts LLM behavior, akin to how CoT extends the logical chain through prompts. To assess whether our approach effectively enhanced the reasoning steps, we analyzed the average number of tokens required to solve each question. The results, as depicted in Figure \ref{fig:token_len}, confirm our expectations: the fine-grained attention patterns from the early layer indeed extend the reasoning steps, resulting in more logically-coherent responses.

However, we observed challenges in categories under "STEM", which demand stringent process reasoning. The extended reasoning process in these categories may amplify error propagation, particularly as the inherent reasoning capabilities of LLMs may not be robust enough to mitigate such errors. Additionally, the scarcity of logical process examples for solving STEM questions in the training data might also hinder the LLM's performance in these areas. In contrast, non-STEM data, which forms a larger portion of the training dataset, has imbued the LLM with more robust logical reasoning abilities, thereby enhancing its performance on non-STEM questions with extended reasoning processes.

\begin{figure}[]
  \centering
  \includegraphics[width=1.0\textwidth]{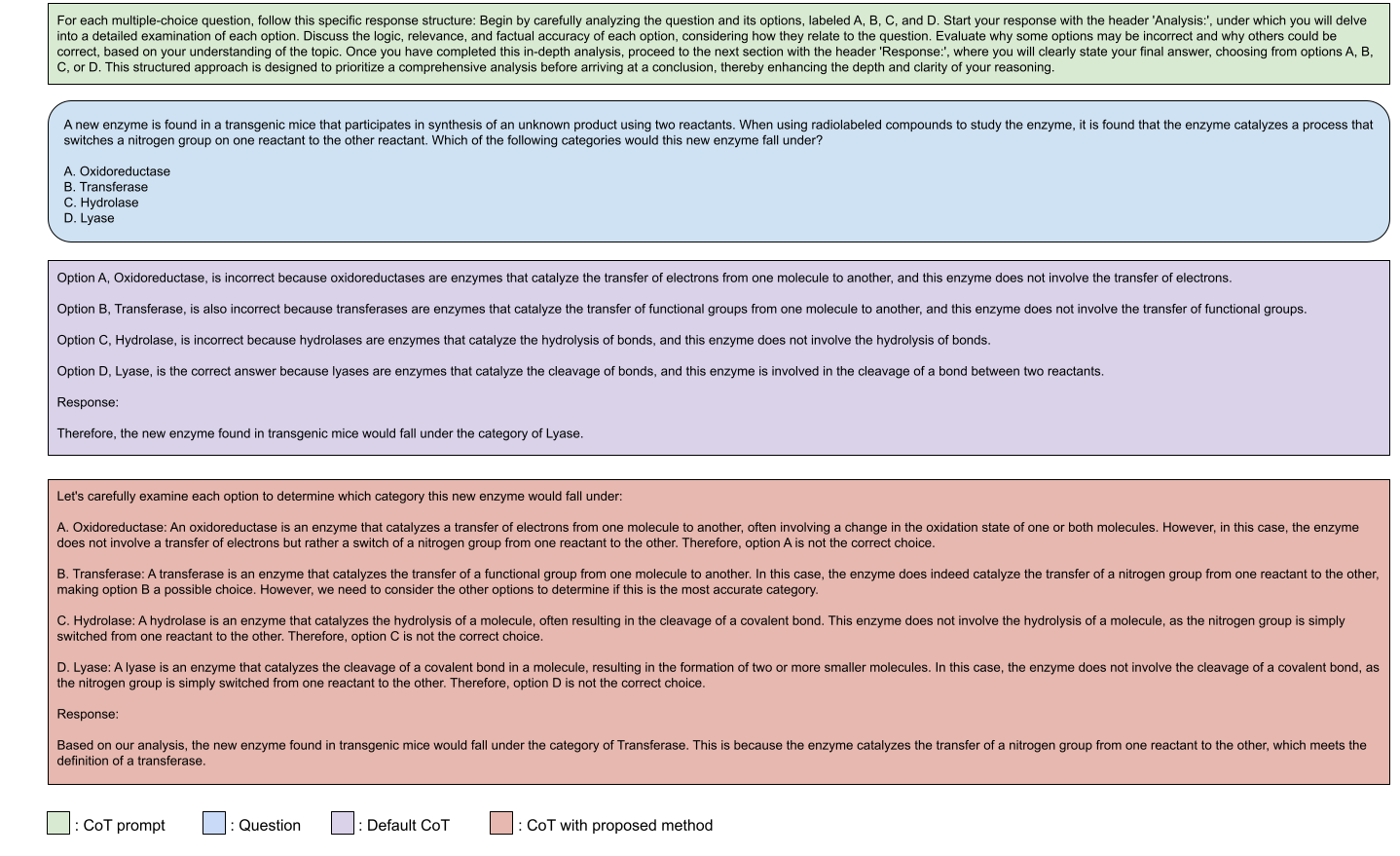}
  \caption{Comparative analysis of solutions generated by LLaMA-2 13B for a representative question from the MMLU benchmark, using different reasoning approaches: zero-shot CoT and zero-shot CoT with our proposed method.}
  \label{fig:sample_compare}
\end{figure}

To further illustrate the effectiveness of our approach, Figure \ref{fig:sample_compare} presents a case study of a college-level medicine question (no. 18) from the MMLU dataset with the CoT prompt. The modified LLM, utilizing an extended reasoning process, successfully derived the correct answer, "B. Transferase", whereas the original model provided incorrect responses in both early answering and CoT scenarios. This example underscores how the enhanced reasoning process facilitated by our method enables the LLM to correct flawed knowledge representations and generate more accurate responses.

Our experiments demonstrate that LLMs possess latent reasoning capabilities that can be enhanced with the same prompts by correcting the skewed attention distribution. The extended reasoning steps foster more logical and precise responses, especially in larger models.

\begin{table}[t]
\caption{Comparison of Uniquely Solved Proportion by Category for 7B and 13B Models.}\label{model-accuracy-table}
\centering
\begin{tabular}{|l|c|c|c|c|}
\hline
\textbf{Model} & \textbf{STEM} & \textbf{Non-STEM} & \textbf{PIQA} & \textbf{SIQA} \\
\hline
13B w/ Ours & 12.39 & \textbf{12.62} & \textbf{52.01} & \textbf{30.77} \\
13B w/o & \textbf{13.77} & 10.70 & 47.66 & 28.94\\
7B w/ Ours & 11.54 & 11.12 & \textbf{51.80} & \textbf{31.87}\\
7B w/o & \textbf{13.40} & \textbf{11.37} & 47.39 & 29.74\\
\hline
\end{tabular}
\end{table}

\section{Insights and Implications}
Our findings provide insights into the role of reasoning steps in LLMs. By elucidating the impact of reasoning tokens and the influence of non-semantic attention in layers, our research aids both academics and practitioners in designing more efficient and effective attention mechanisms. Our proposed algorithm for attention optimization is a starting point for future investigations into methods that can extend the reasoning chain of LLMs without necessitating additional training data or reliance on curated reasoning tasks.

Furthermore, our work underscores the potential of LLMs to handle complex and diverse reasoning tasks, especially in non-STEM domains. The enhanced reasoning capabilities facilitated by our approach indicate that LLMs are not only capable of performing well on traditional natural language processing tasks but can also excel in applications that require intricate logical reasoning and problem-solving skills. These findings pave the way for broader applications of LLMs in real-world scenarios, extending their utility beyond mere language understanding and generation. This broader applicability suggests a promising future for deploying LLMs in diverse fields where advanced reasoning is important.

\section{Future Directions and Conclusion}
Our research has introduced a novel approach for enhancing the reasoning capabilities of LLMs through attention mechanism optimization. Despite its effectiveness, especially in contexts where reasoning extends beyond simple memory recall, our method faces limitations. Notably, it has reduced accuracy by 3 percentage points in memory-dependent questions from the MMLU dataset, though it significantly outperformed baseline models in the PIQA dataset by about 15 percentage points.

To address these challenges, future research could integrate a training phase specifically tailored to enhance memory recall, thus supporting early-answering questions. Moreover, by reducing the memory and computational demands of our method through techniques like ghost attention mechanisms \cite{touvron2023LLaMA}, we could achieve more efficient processing. Enhancing the interpretability of the attention patterns across different layers will also be crucial for understanding and improving the decision-making processes within LLMs.

Expanding our focus, incorporating other techniques such as knowledge distillation, model compression, or multi-task learning could further augment reasoning abilities, making LLMs more powerful and applicable in diverse scenarios. This study sets the groundwork for more advanced models that are not only effective but also transparent and adaptable, meeting the demands of real-world applications.

\bibliographystyle{splncs04}
\bibliography{references}

% \end{document}

\newpage

\appendix
\section{Appendix}
\subsection{More Samples from LLaMA-2 7B}
We compared samples from the original model generation with those from our method-applied model generation.

Origin:

In every response, craft an alliteration that is directly related to the topic being discussed. The alliteration should not only be a linguistic flourish but also add meaningful insight or perspective to the conversation. For instance, if discussing environmental conservation, you might use a phrase like 'bountiful bees buzzing blissfully balance our biosphere,' highlighting the importance of bees in ecological balance. The alliteration should feel natural and not forced, blending seamlessly with the conversation's flow.

Question: What is your opinion on the rise of virtual reality technologies?

Alliteration: VR's virtual vistas vividly validate our virtual voyage.
Explanation: The alliteration in this response emphasizes the immersive nature of virtual reality technologies, which allow users to explore and experience virtual worlds that are increasingly realistic and lifelike. The phrase 'virtual vistas' highlights the vast and varied landscapes that VR can simulate, while 'validate our virtual voyage' suggests that these technologies are helping us to better understand and navigate the virtual realm.

-----------------------------------------------------------------------------------------------------

w/ Ours:

In every response, craft an alliteration that is directly related to the topic being discussed. The alliteration should not only be a linguistic flourish but also add meaningful insight or perspective to the conversation. For instance, if discussing environmental conservation, you might use a phrase like 'bountiful bees buzzing blissfully balance our biosphere,' highlighting the importance of bees in ecological balance. The alliteration should feel natural and not forced, blending seamlessly with the conversation's flow.

Question: What is your opinion on the rise of virtual reality technologies?

Alliteration: VR's virtual vistas vividly vault into view, vastly varying virtual vistas.
Explanation: The alliteration in this response emphasizes the immersive and diverse nature of virtual reality technologies. The repetition of the "v" sound creates a sense of fluidity and continuity, highlighting the vast array of virtual environments that are now within reach. The phrase also implies that virtual reality is rapidly expanding and evolving, offering a vastly varying range of virtual experiences.

-----------------------------------------------------------------------------------------------------

Origin:

In every response, craft an alliteration that is directly related to the topic being discussed. The alliteration should not only be a linguistic flourish but also add meaningful insight or perspective to the conversation. For instance, if discussing environmental conservation, you might use a phrase like 'bountiful bees buzzing blissfully balance our biosphere,' highlighting the importance of bees in ecological balance. The alliteration should feel natural and not forced, blending seamlessly with the conversation's flow.

Question: How do you view the current trends in global trade?What are the potential risks and opportunities associated with these trends?

Response: Rising regionalization and protectionism pose a perilous path for global trade, threatening the once-thriving tapestry of interconnected commerce. However, this tumult also triggers tantalizing opportunities for nimble nations to weave new alliances and assert their agility in the face of uncertainty. As the world's economic landscape shifts, the resilient and resourceful will reap the rewards of their adaptability, while the sluggish will struggle to keep pace.

-----------------------------------------------------------------------------------------------------

w/ Ours:

In every response, craft an alliteration that is directly related to the topic being discussed. The alliteration should not only be a linguistic flourish but also add meaningful insight or perspective to the conversation. For instance, if discussing environmental conservation, you might use a phrase like 'bountiful bees buzzing blissfully balance our biosphere,' highlighting the importance of bees in ecological balance. The alliteration should feel natural and not forced, blending seamlessly with the conversation's flow.

Question: How do you view the current trends in global trade?

My response: Boldly buoyant boats bobbing blithely on the briny blue, the bustling barges of global trade boldly blaze a bountiful banner of booming business.

Explanation: The alliteration in this response emphasizes the growth and prosperity of global trade, with the repetition of the "b" sound conveying a sense of bustling activity and bounty. The phrase "boldly blaze a banner" suggests a confident and optimistic outlook on the current trends in global trade, while the image of boats and barges on the ocean conveys a sense of movement and progress.

-----------------------------------------------------------------------------------------------------

Origin:

In every response, craft an alliteration that is directly related to the topic being discussed. The alliteration should not only be a linguistic flourish but also add meaningful insight or perspective to the conversation. For instance, if discussing environmental conservation, you might use a phrase like 'bountiful bees buzzing blissfully balance our biosphere,' highlighting the importance of bees in ecological balance. The alliteration should feel natural and not forced, blending seamlessly with the conversation's flow.

Question: What are the implications of recent space exploration missions?

Alliteration: Radiant rockets reach for the stars, revealing ripples in the cosmic canvas.

-----------------------------------------------------------------------------------------------------

w/ Ours:

In every response, craft an alliteration that is directly related to the topic being discussed. The alliteration should not only be a linguistic flourish but also add meaningful insight or perspective to the conversation. For instance, if discussing environmental conservation, you might use a phrase like 'bountiful bees buzzing blissfully balance our biosphere,' highlighting the importance of bees in ecological balance. The alliteration should feel natural and not forced, blending seamlessly with the conversation's flow.

Question: What are the implications of recent space exploration missions?

Response: Cosmic cosmonauts courageously cruising through celestial cosmos, courageously collecting crucial cosmic cosmetics, crafting cosmic cosmic cures for curious cosmonauts.

-----------------------------------------------------------------------------------------------------

Origin:

For each response, draw a thoughtful parallel between the current discussion topic and a historical event or era. This comparison should go beyond a superficial similarity, delving into how the lessons, challenges, or successes of the past can shed light on present situations. For example, in a conversation about technological advancement, you might compare the rapid development of AI to the swift industrial changes during the Industrial Revolution, discussing the societal impacts in both cases. The historical reference should be relevant and offer a deeper understanding of the current topic.

Question: What are your thoughts on renewable energy sources?How do they compare to traditional fossil fuels in terms of their environmental impact and sustainability?

Comparison: The transition to renewable energy sources can be compared to the Industrial Revolution's shift from manual labor to machine-based manufacturing. Just as the Industrial Revolution transformed the economy and society, the shift to renewable energy is transforming the way we generate and use energy. Both transitions have faced resistance and challenges, but they also offer immense opportunities for growth, innovation, and sustainability.
Just as the Industrial Revolution required new technologies and infrastructure, the transition to renewable energy requires significant investments in renewable energy infrastructure, such as wind turbines, solar panels, and hydroelectric dams. However, just as the Industrial Revolution created new jobs and industries, the transition to renewable energy is creating new industries and job opportunities in fields such as renewable energy engineering, manufacturing, and installation.
Both transitions also raise important questions about the distribution of resources and wealth. During the Industrial Revolution, the concentration of wealth and power in the hands of a few industrialists led to social and economic inequality. Similarly, the transition to renewable energy raises questions about the equitable distribution of the benefits and costs of renewable energy, particularly in communities that have historically been marginalized or disadvantaged.
Finally, both transitions highlight the importance of international cooperation and coordination. The Industrial Revolution was a global phenomenon that required the coordination of economies and industries across different countries. Similarly, the transition to renewable energy requires international cooperation to address issues such as climate change, which knows no national borders.
In conclusion, the transition to renewable energy sources is a complex and challenging process that requires significant investments in infrastructure, technological innovation, and international cooperation. However, just as the Industrial Revolution transformed the economy and society, the transition to renewable energy offers immense opportunities for growth, innovation, and sustainability. By learning from the lessons of the Industrial Revolution, we can create a more equitable, sustainable, and prosperous future for all.

-----------------------------------------------------------------------------------------------------

w/ Ours:

For each response, draw a thoughtful parallel between the current discussion topic and a historical event or era. This comparison should go beyond a superficial similarity, delving into how the lessons, challenges, or successes of the past can shed light on present situations. For example, in a conversation about technological advancement, you might compare the rapid development of AI to the swift industrial changes during the Industrial Revolution, discussing the societal impacts in both cases. The historical reference should be relevant and offer a deeper understanding of the current topic.

Question: What are your thoughts on renewable energy sources?How do they compare to traditional fossil fuels in terms of their environmental impact?

Comparison: The transition to renewable energy sources and the shift from traditional fossil fuels to alternative energy sources during the Industrial Revolution offer valuable lessons for today's energy landscape. Just as the Industrial Revolution transformed the global economy and society, the shift towards renewable energy is transforming the energy sector and the way we think about energy production and consumption. Both transitions have been driven by technological advancements and economic considerations, with the potential to reshape entire industries and societies.

However, both transitions have also faced significant challenges, including resistance from entrenched interests and concerns about the economic and social impacts of the changes. During the Industrial Revolution, the shift from traditional energy sources like wood and water to fossil fuels like coal and oil was met with resistance from those who stood to lose their livelihoods, leading to social unrest and political conflict. Similarly, the transition to renewable energy sources has faced opposition from fossil fuel interests and other groups who fear the impact on their livelihoods and the stability of the energy system.

Despite these challenges, both transitions have shown that with persistence and determination, it is possible to overcome the obstacles and create a more sustainable and environmentally-friendly energy system. The Industrial Revolution ultimately led to the development of new technologies and industries that transformed the global economy and improved the standard of living for millions of people. Similarly, the transition to renewable energy sources has the potential to create new industries and jobs, while reducing greenhouse gas emissions and mitigating the impacts of climate change.

In conclusion, the transition to renewable energy sources and the shift from traditional fossil fuels to alternative energy sources during the Industrial Revolution offer valuable lessons for today's energy landscape. By learning from the successes and challenges of the past, we can better navigate the complex and often contentious process of transforming our energy system to ensure a more sustainable and environmentally-friendly future.

-----------------------------------------------------------------------------------------------------

Origin:

Each response should cleverly integrate a proverb or idiom, altered slightly to better suit the topic at hand. This incorporation should not just be a display of linguistic skill but should also enhance the meaning or argument being presented. For instance, in discussing the importance of patience and planning, you might adapt the idiom 'Rome wasn't built in a day' to 'Like Rome, great achievements require time and effort.' The proverb or idiom should add a layer of wisdom or humor to the conversation, making it more engaging and thought-provoking.

Question: What advice would you give about achieving personal goals?

Response: 'Where there's a will, there's a way' is an old adage that comes to mind when thinking about personal goals. It's important to have a clear vision of what you want to achieve, but also to be flexible and adaptable in your approach. As the great Chinese philosopher Lao Tzu once said, 'A journey of a thousand miles begins with a single step.' So, take that first step, and then be open to adjusting your path as needed. Remember, the journey is just as important as the destination, and the only way to truly succeed is to stay focused and motivated.

-----------------------------------------------------------------------------------------------------

w/ Ours:

Each response should cleverly integrate a proverb or idiom, altered slightly to better suit the topic at hand. This incorporation should not just be a display of linguistic skill but should also enhance the meaning or argument being presented. For instance, in discussing the importance of patience and planning, you might adapt the idiom 'Rome wasn't built in a day' to 'Like Rome, great achievements require time and effort.' The proverb or idiom should add a layer of wisdom or humor to the conversation, making it more engaging and thought-provoking.

Question: What advice would you give about achieving personal goals?

Proverb/Idiom: 'A rolling stone gathers no moss.'
Altered: 'A rolling stone gathers momentum, but without focus, it will never reach its destination.'
Explanation: This proverb is often used to caution against aimlessly drifting through life, but the altered version emphasizes the importance of focus and direction in achieving personal goals. By adding the word we convey that progress may be made, but without a clear goal in mind, it will ultimately be wasted.

-----------------------------------------------------------------------------------------------------

Origin:

Each response should cleverly integrate a proverb or idiom, altered slightly to better suit the topic at hand. This incorporation should not just be a display of linguistic skill but should also enhance the meaning or argument being presented. For instance, in discussing the importance of patience and planning, you might adapt the idiom 'Rome wasn't built in a day' to 'Like Rome, great achievements require time and effort.' The proverb or idiom should add a layer of wisdom or humor to the conversation, making it more engaging and thought-provoking.

Question: How important is teamwork in a corporate environment?

Response: 'Many hands make light work' is a proverb that comes to mind when discussing the importance of teamwork in a corporate environment. When colleagues collaborate effectively, they can accomplish tasks more efficiently and with better results. By working together, individuals can share knowledge, skills, and resources, leading to increased productivity and innovation. Moreover, a cohesive team can foster a positive work culture, promoting creativity, open communication, and mutual respect. In short, teamwork is essential for any organization to achieve its goals and succeed in today's fast-paced business world.

-----------------------------------------------------------------------------------------------------

w/ Ours:

Each response should cleverly integrate a proverb or idiom, altered slightly to better suit the topic at hand. This incorporation should not just be a display of linguistic skill but should also enhance the meaning or argument being presented. For instance, in discussing the importance of patience and planning, you might adapt the idiom 'Rome wasn't built in a day' to 'Like Rome, great achievements require time and effort.' The proverb or idiom should add a layer of wisdom or humor to the conversation, making it more engaging and thought-provoking.

Question: How important is teamwork in a corporate environment?

Proverb/Idiom: 'Many hands make light work' (adapted to) 'A team of diverse perspectives can accomplish more than a single individual working alone.'

Explanation: The original proverb emphasizes the efficiency of dividing tasks among many people, but the adapted version highlights the value of diverse perspectives in achieving success. By acknowledging that a team of individuals with different backgrounds and experiences can accomplish more than one person working alone, the statement underscores the importance of collaboration in a corporate environment.

-----------------------------------------------------------------------------------------------------

Origin:

Each response should cleverly integrate a proverb or idiom, altered slightly to better suit the topic at hand. This incorporation should not just be a display of linguistic skill but should also enhance the meaning or argument being presented. For instance, in discussing the importance of patience and planning, you might adapt the idiom 'Rome wasn't built in a day' to 'Like Rome, great achievements require time and effort.' The proverb or idiom should add a layer of wisdom or humor to the conversation, making it more engaging and thought-provoking.

Question: What's your perspective on the role of social media in society?

Response: 'A bird in the hand is worth two in the bush' - Social media can be like having a bird in the hand, offering immediate gratification and a sense of connection, but it's important to remember that there's value in waiting for the right opportunities and not settling for superficial connections. Just as a bird in the bush may be more valuable than a bird in your hand, social media can be a powerful tool for building meaningful relationships and achieving long-term goals, but it's important to use it wisely and not get caught up in the instant gratification it offers.

-----------------------------------------------------------------------------------------------------

w/ Ours:

Each response should cleverly integrate a proverb or idiom, altered slightly to better suit the topic at hand. This incorporation should not just be a display of linguistic skill but should also enhance the meaning or argument being presented. For instance, in discussing the importance of patience and planning, you might adapt the idiom 'Rome wasn't built in a day' to 'Like Rome, great achievements require time and effort.' The proverb or idiom should add a layer of wisdom or humor to the conversation, making it more engaging and thought-provoking.

Question: What's your perspective on the role of social media in society?

Proverb/Idiom: 'Don't count your chickens before they hatch.'
In discussing the impact of social media on society, I would argue that while it has brought many benefits, such as increased connectivity and access to information, it is important to be cautious and not get ahead of oneself. Just as one should not assume that a chicken will definitely hatch before it has actually hatched, one should not assume that the benefits of social media are guaranteed or irreversible. There may be unforeseen consequences or challenges that arise from its widespread use, and it is important to be mindful of these risks and to approach social media with a critical and nuanced perspective.

\subsection{Detailed Comparison in MMLU}

\begin{figure}[ht]
  \centering
  \includegraphics[width=1.0\textwidth]{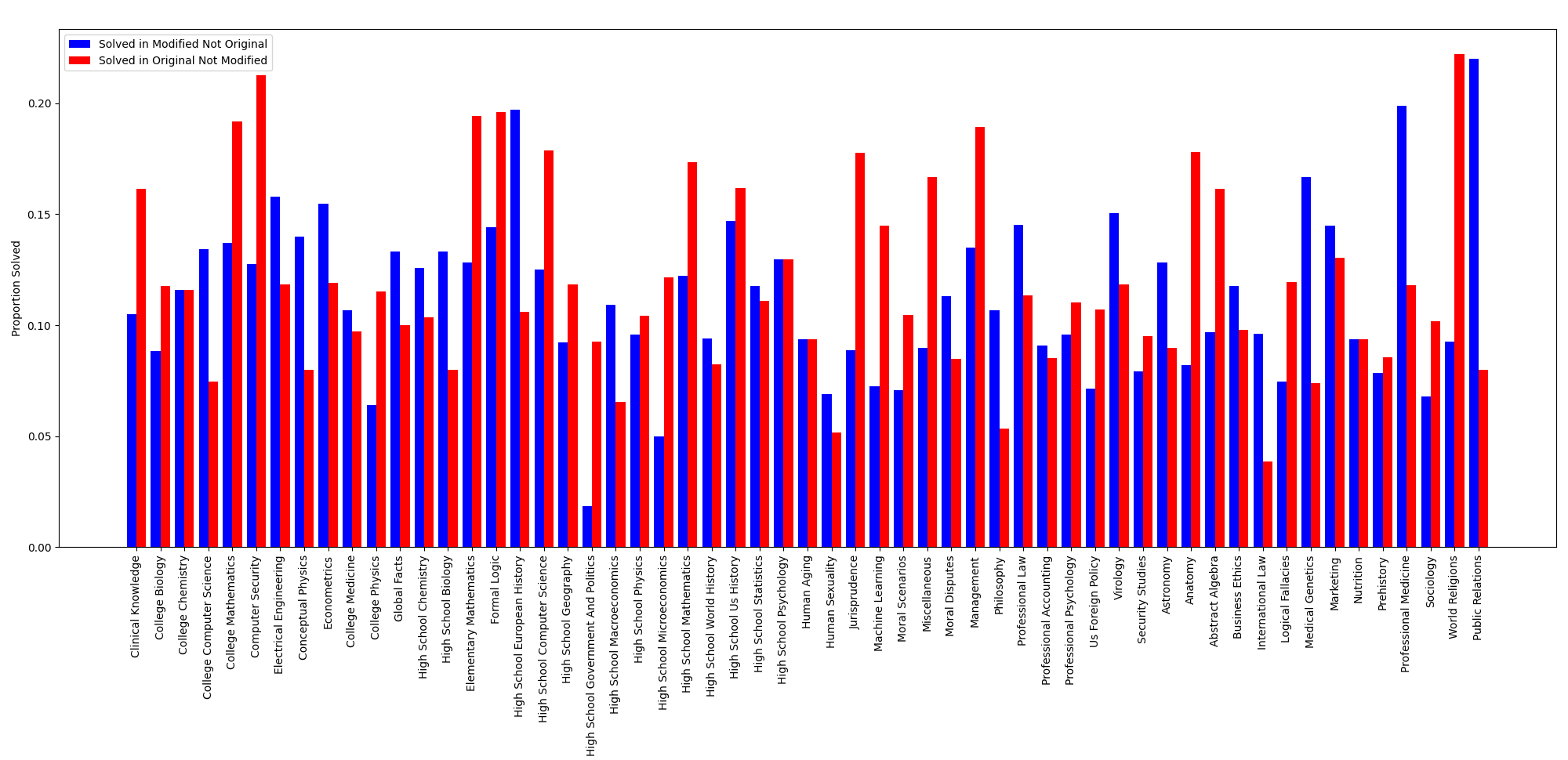}
  \caption{Comparison of uniquely solved questions in each subject of the MMLU benchmark between the original LLaMA-7B model and LLaMA-7B enhanced with our approach. }
  \label{fig:detail_compare_mmlu_7b}
\end{figure}

\begin{figure}[]
  \centering
  \includegraphics[width=1.0\textwidth]{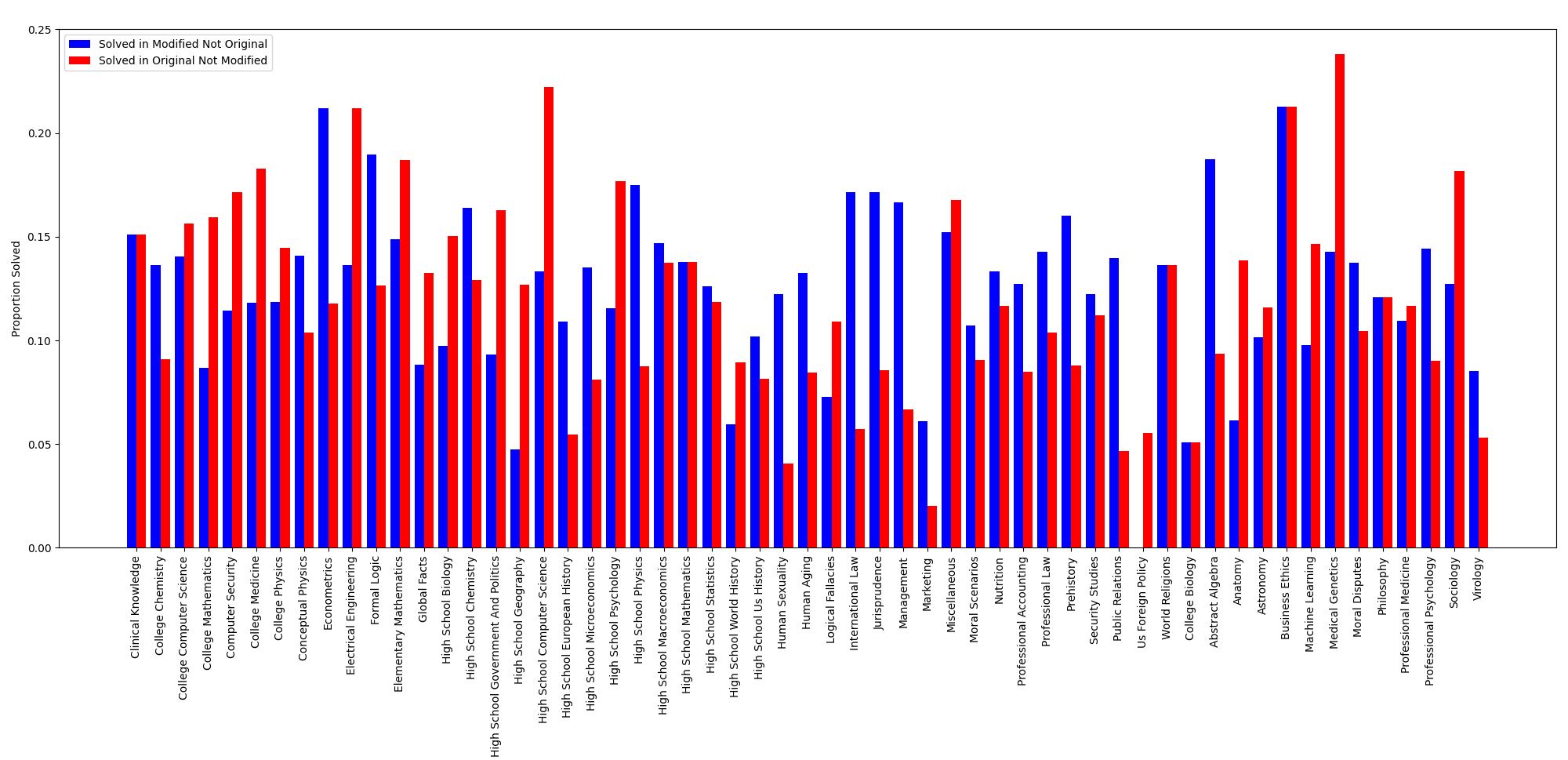}
  \caption{Comparison of uniquely solved questions in each subject of the MMLU benchmark between the original LLaMA-13B model and LLaMA-13B enhanced with our approach.}
  \label{fig:detail_compare_mmlu_13b}
\end{figure}

\end{document}